\documentclass[runningheads]{llncs}
\usepackage{graphicx}
\usepackage{times}
\usepackage{latexsym}

\usepackage[T1]{fontenc}
\usepackage[utf8]{inputenc}

\usepackage{microtype}

\usepackage{helvet} % DO NOT CHANGE THIS
\usepackage{courier}  % DO NOT CHANGE THIS
\usepackage{url}  % DO NOT CHANGE THIS
\usepackage{graphicx} % DO NOT CHANGE THIS
\usepackage{wrapfig}
\usepackage[utf8]{inputenc}
\usepackage{subfigure}
\usepackage{booktabs}
\usepackage{amsmath}
\usepackage{amsfonts}
\usepackage{multirow}
\usepackage{color}
\usepackage[lined,boxed,commentsnumbered, ruled,linesnumbered]{algorithm2e}
\usepackage{enumitem}
\usepackage{microtype}
\usepackage{tabularx}
\usepackage{arydshln}
\usepackage{chngpage}
\usepackage{array}
\usepackage{hyperref}

\usepackage{microtype}
\usepackage{makecell}
\usepackage{tabularx}
\usepackage{booktabs}

\usepackage{marvosym}

\begin{document}
	\title{EviDR: Evidence-Emphasized Discrete Reasoning for \\ Reasoning Machine Reading Comprehension}
	% \title{Reasoning Machine Reading Comprehension with Evidence}
	%
	\titlerunning{EviDR}
	% \orcidID{0000-0002-5549-5130}
	\author{Yongwei Zhou\inst{1} \and
		Junwei Bao\inst{2} \and
		Haipeng Sun\inst{2} \and
		Jiahui Liang\inst{2} \and \\
		Youzheng Wu\inst{2} \and
		Xiaodong He\inst{2} \and
		Bowen Zhou\inst{2} \and
		Tiejun Zhao\inst{1}\thanks{Corresponding author. Work done during the first author’s internship at JD AI Research.
	}}

	\authorrunning{Yongwei Zhou et al.}
	\institute{Harbin Institute of Technology, Harbin, China \and
		JD AI Research, Beijing, China \\
		\tt ywzhou@hit-mtlab.net \;\; \tt tjzhao@hit.edu.cn \\
		\tt \{baojunwei,sunhaipeng6,liangjiahui14\}@jd.com \\
		\tt \{wuyouzheng1,xiaodong.he,bowen.zhou\}@jd.com \\
	}
	
	\maketitle              % typeset the header of the contribution

	\begin{abstract}
		Reasoning machine reading comprehension (R-MRC) aims to answer complex questions that require discrete reasoning based on text.
		To support discrete reasoning, evidence, typically the concise textual fragments that describe question-related facts,  including topic entities and attribute values, are crucial clues from question to answer.
		However, previous end-to-end methods that achieve state-of-the-art performance rarely solve the problem by paying enough emphasis on the modeling of evidence, missing the opportunity to further improve the model's reasoning ability for R-MRC.
		To alleviate the above issue, in this paper, we propose an \underline{Evi}dence-emphasized \underline{D}iscrete \underline{R}easoning approach (\textbf{EviDR}), in which sentence and clause level evidence is first detected based on distant supervision, and then used to drive a reasoning module implemented with a relational heterogeneous graph convolutional network to derive answers.
		Extensive experiments are conducted on DROP (discrete reasoning over paragraphs) dataset, and the results demonstrate the effectiveness of our proposed approach.
		In addition, qualitative analysis verifies the capability of the proposed evidence-emphasized discrete reasoning for R-MRC. \footnote{Code is released at \url{https://github.com/JD-AI-Research-NLP/EviDR}}.
		
	\end{abstract}
	\section{Introduction}
	Machine Reading Comprehension (MRC) aims to answer questions based on text, which has recently been widely explored and achieved remarkable progress that some methods have approached and even outperformed humans~\cite{clark2020electra,glass2020span,lan2019albert,li2020hopretriever}.
	Most of these studies focus on MRC datasets which have been released around span extraction, e.g., SQUAD~\cite{rajpurkar2016squad}, conversational state tracking, e.g., CoQA~\cite{reddy2019coqa} and QuAC~\cite{choi2018quac}, passage retrieval, e.g., HotpotQA~\cite{yang2018hotpotqa}, multi-choice selection, e.g., RACE~\cite{lai2017race} and CommonsenseQA~\cite{talmor2019commonsenseqa}, and answer generation, e.g., MS-MARCO~\cite{nguyen2016ms} and DuReader~\cite{he2018dureader}.
	However,  the reasoning capability of MRC models, which is especially crucial for the comprehension of sports news, scientific reports, and financial news where much arithmetic computation is required, has rarely been evaluated on these datasets.
	\begin{table*}[!htbp]
		% \vspace{-0.5cm}
		\centering
		\caption{\label{example}An Example of question-answer pairs along with document from the DROP dataset. Sentence level evidences are in {\textcolor{blue}{blue}}, and clause level evidences are further in {\textcolor{blue}{\textbf{bold}}}. \textbf{Ques} means question and \textbf{Ans} indicates answer.}
		\begin{tabularx}{\textwidth}{lp{0.92\textwidth}}
			\hline
			\textbf{Text:} & \textcolor[RGB]{0,0,0}{As of the census of 2010, there were 31,894 people, 13,324 households, and 8,201 families residing in the city. The population density was 1,851.1 inhabitants per square mile (714.7/km$^2$). There were 14,057 housing units at an average density of 815.8 per square mile (315.0/km$^2$). {\textcolor{blue}{\textbf{The racial makeup of the city was \underline{93.9\%} \underline{White} (U.S. Census)}, 0.3\% African American (U.S. Census), 1.7\% Native American (U.S. Census), {\textbf{\underline{0.8\%} \underline{Asian} (U.S. Census)}}, 0.1\% Race (U.S. Census), 0.7\% from Race (U.S. Census), and 2.4\% from two or more races.}} Hispanic (U.S. Census) or Latino (U.S. Census) of any race were 2.8\% of the population.} \\
			\textbf{Ques:} & {\textcolor{blue}{How many more percentage of the population had a racial make-up of \underline{White} than \underline{Asian}?}}\\
			\textbf{Ans:} & 93.1\\
			\hline
		\end{tabularx}%
		\vspace{-0.5cm}
	\end{table*}%

	To complement the above shortage of research on reasoning machine reading comprehension (R-MRC), DROP~\cite{dua2019drop} has recently been proposed for tracking complex questions that require discrete reasoning over text.
	Table~\ref{example} shows an example of DROP dataset, where the ground truth answer $93.1$ to the question should be derived by arithmetic computation $93.9-0.8 = 93.1$, especially referring to the evidences, textual fragments in blue, including topic entities, i.e., \underline{White} and \underline{Asian}, and attribute values, i.e., \underline{93.9\%} and \underline{0.8\%}.
	As a preliminary attempt toward the task, NAQANET~\cite{dua2019drop} is proposed as a number-aware framework to deal with questions with multi-predictors corresponding to different answer types,  including span, count, and addition/subtraction.
	Based on NAQANET, NumNet~\cite{ran2019numnet} and QDGAT~\cite{chen2020question} perform reasoning over a heuristic graph including numerical values or additional entities with graph neural network to enhance the reasoning abilities.
	Although these end-to-end methods achieve state-of-the-art performance, they have rarely placed enough emphasis on explicitly modeling of evidence that is typically crucial clues from question to answer, which miss the opportunity to further improve reasoning ability for R-MRC.
	
	% which are mainly based on multi-head predictors framework, which means proposing different answer predictors upon a shared encoder for corresponding answer type.
	% However, like NumNet or QDGAT, introducing a graph reasoning module integrate some structured knowledge into the representation of sequence, a real reasoning process bridge the question and answer remains to be a black box. 
	% 1. 增加推理模块只是增加了额外的知识，问题和答案之间缺乏一个真正的推理过程
	% 2.
	
	In this paper, we propose to address the R-MRC problem with \underline{Evi}dence-emphasized \underline{D}iscrete \underline{R}easoning (\textbf{EviDR}).
	First, evidence is pinpointed with an evidence detector finetuned on a pre-trained language model via distant supervision.
	In detail, the evidence detector is trained to judge whether a textual fragment is evidence or not, where the distant supervision signal is obtained under one-shot heuristic rules without human annotation.
	We adopt multi-granularity evidence,  including sentence-level and clause level,  as a trade-off between recall and precision for evidence detection.
	Then, information about evidence, including evidence representations and evidence pinpointing distribution over text, are used to drive a reasoning module to derive answers.
	Specifically, the reasoning module is implemented with a relational heterogeneous graph convolutional network (RH-GCN) upon the same encoder to explicitly propagate and emphasize the information of evidence.
	The heterogeneous graph is constructed based on sentence-level and clause-level nodes linked with different edges and updated with evidence pinpointing distribution as weights.
	In general, our model is jointly trained with multi-tasks,  including evidence detection and reading comprehension.
	
	%We conduct experiments on the DROP dataset to show the effectiveness of our approach.
	Experiment results on the DROP dataset show that our approach achieves significant improvements compared with a strong baseline built upon a pre-trained language model without evidence modeling, and similar and even better results compared to the state-of-the-art model, i.e., QDGAT.
	%, re-implemented by us.
	Besides, the ablation study verifies the effectiveness of the distant supervision of evidence detection and the proposed evidence-emphasized discrete reasoning module with RH-GCN.
	Moreover, qualitative analysis verifies the reasoning ability of our proposed approach for R-MRC.
	
	The contributions of the paper include the following three aspects.
	(1) We propose an evidence detector to explicitly pinpoint multi-granularity evidence as clues, which is learned via distant supervision.
	(2) We propose an evidence-emphasized discrete reasoning network with a relational heterogeneous graph convolutional network, which enhances the reasoning ability of R-MRC models.
	(3) We conduct extensive experiments and analysis on DROP, proving the effectiveness of the approach, and verifying the capability of the evidence-emphasized discrete reasoning for R-MRC.

	\section{Related Work}
	%\subsection{Benchmarks and Models}
	% Recently, MRC has drawn wide attention, many popular datasets have been released to evaluate the capabilities of MRC models, such as SQuAD \cite{rajpurkar2016squad}, QuAC \cite{choi2018quac}, RACE \cite{lai2017race} and so on. A number of MRC models have been proposed for these benchmarks and achieved remarkable performance on these benchmarks, such as Retro-Reader \cite{zhang2020retrospective}, QANET \cite{yu2018qanet} , BiDAF \cite{seo2016bidirectional} and massive models based on pre-trained language models, e.g. BERT \cite{devlin2018bert}, RoBERTa \cite{liu2019roberta}, ELECTRA \cite{clark2020electra} and ALBERT \cite{lan2019albert}. However, these models lack the discrete reasoning ability for R-MRC task.
	
	Recently, two lines of approaches have been proposed for the R-MRC task. The first is based on semantic parsing. Dua \emph{et al}.~\cite{dua2019drop} converted the unstructured text into a table and adopted a grammar-constrained semantic parsing model named KDG to answer the question over the table \cite{krishnamurthy2017neural}. Chen \emph{et al}.~\cite{chen2019neural} proposed a generative model NeRd, which is composed of a reader and a programmer. They are responsible for encoding questions and passages into vector representation and generate grammatical programs, respectively. Gupta \emph{et al}.~\cite{gupta2019neural} learned to parse compositional questions as executable programs where each atomic program is a learnable neural module. However, the model only adapted to questions with predefined templates. The second is based on neural end-to-end methods. As a preliminary attempt toward the task, Dua \emph{et al}.~\cite{dua2019drop} proposed a number-aware framework named NAQANET to produce three different answer types with various predictors, including a span, count, and arithmetic expression. To aggregate relative magnitude relation between two numbers, NumNet~\cite{ran2019numnet} was proposed to perform multi-step reasoning over a number comparison graph. Chen \emph{et al}.~\cite{chen2020question} proposed QDGAT based on a heterogeneous graph, which aggregates both entity and number nodes information. GenBERT \cite{geva2020injecting} was proposed to inject the discrete reasoning abilities into BERT by generating numerical data. Compared to these existing methods, our proposed method focuses on placing enough emphasis on the evidence modeling to enhance discrete reasoning ability for the R-MRC model.
	
	\section{Methodology}
	%\subsection{Task Formulation}
	
	\begin{figure*}[tb]
		\centering
		\includegraphics[width=4.8 in]{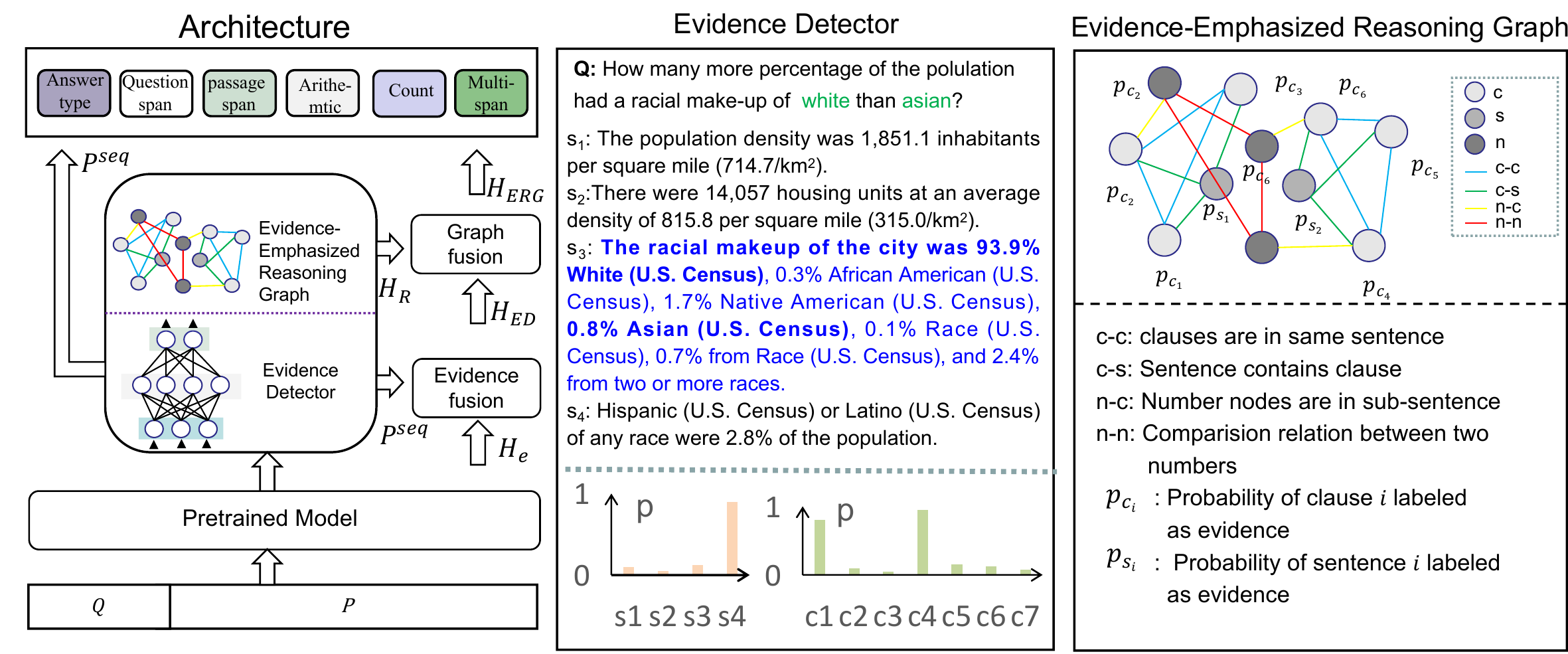}
		\caption{\label{model} An illustration of EviDR architecture. It consists of an encoder, an evidence detector, an evidence-emphasized reasoning graph, and a prediction module. Multi-granularity evidence, including sentence-level and clause-level, is pinpointed through the evidence detection. The evidence-emphasized reasoning module performs multi-step reasoning over a heterogeneous graph in which nodes are weighted by evidence, including sentence nodes, clause nodes, and number nodes. The prediction module supports five kinds of answer types, i.e., question span, passage span, arithmetic expression, count, and multi-spans.}
		\vspace{-0.5cm}
	\end{figure*}

	MRC aims to predict an answer $A$ with the maximum probability $P$ according to the given question $Q$ and passage text $P$:
	\begin{equation}
		A = \arg \max_{\hat{A} \in \Omega} P(\hat{A}|P,Q).
	\end{equation}
	Compared to traditional span extraction MRC, in the R-MRC task, the answer $A$ can not only be spans (single or multiple spans) from the question or passage but also a number obtained by arithmetic computations with some numbers in the context. For questions with a span answer, discrete reasoning is also required in the R-MRC task, such as sort and comparison. In this paper, we explicitly model evidence over text to enhance the reasoning ability of R-MRC systems.

	%\subsection{Framework}
	The framework of our  proposed model \textbf{EviDR} 
	is briefly described in Figure \ref{model}, which is mainly composed of four components, i.e.,  an encoder, an evidence detector, an evidence-emphasized reasoning module, and a prediction module. The encoder is responsible for semantic comprehension for the context of question and passage, commonly implemented by a pre-trained language model at present. Upon the encoder, an evidence detector learns to determine whether the current fragment supports answer prediction.  We construct an evidence-emphasized reasoning graph based on the detector result and integrate knowledge from pieces of evidence through a graph convolution network. Finally, we leverage evidence to guide the answer prediction.

	\subsection{Encoding Module}
	Without loss of generality, we employ pre-trained language model (PLM) as the backbone architecture to encode context of question and passage, which takes concatenation of $\texttt{[CLS]}$, tokens in question, $\texttt{[SEP]}$, tokens in passage and $\texttt{[SEP]}$ as textual inputs, and  the output representation is denoted as:
	\begin{equation}
		\mathbf{H}_e = [\mathbf{H}^{Q}_e; \mathbf{H}^P_e] = \mathtt{PLM}(\mathbf{Q},\mathbf{P}).
	\end{equation}
	% $$ \mathbf{H}^{Q}, \mathbf{H}^P = \mathtt{ELECTRA}(\mathbf{Q},\mathbf{P})$$
	where $\mathbf{H}^Q_e \in \mathbb{R}^{ L_{q}\times d_{h}}$ and $\mathbf{H}^P_e\in \mathbb{R}^{ L_{p}\times d_{h}}$ are the output representations of question and passage, respectively.  $L_q$ and $L_p$ are the length of question and passage tokens, respectively. $d_h$ is the hidden size. 
	
	\subsection{Evidence Detector}
	\label{sec:evidence_retriever}
	\paragraph{Evidence Detector} Exactly as supporting fact prediction task in HotpotQA \cite{yang2018hotpotqa} and documents retrieval procedure in open domain QA task, evidence plays a crucial role in the MRC task. Therefore, we additionally introduce an evidence detector, which is responsible for discriminating whether each fragment can act as evidence to support answer prediction or not. %We formulate the evidence detector as a 0/1 classification task and implement it with a $\mathtt{FFN}(\cdot)$, a feed-forward network which consists of two linear projections with a GeLU activation \cite{} followed by a layer normalization \cite{} in between. 
	
	Specifically, the evidence detector takes the representation of question $\mathbf{S}^{Q}$ and each fragment in passage $\mathbf{S}_k^{P}$ as input features $(k=1,2,...,m)$ and output the probability distribution of being identified as evidence to support answer prediction through a feed-forward network $\mathtt{FFN}(\cdot)$ as follows:
	\begin{equation}
		\begin{split}
			\mathbf{P}_{k} = \mathtt{softmax}(\mathtt{FFN}(\mathbf{S}^{Q}, \mathbf{S}_{k}^{P})), \;\;
			\mathbf{S}^{Q}=\mathbf{\beta}^{Q}\mathbf{H}^{Q}_e, \;\;\mathbf{\beta}^{Q} = \mathtt{softmax}(\mathbf{H}^{Q}_e\mathbf{W}),\\
		\end{split}
	\end{equation}
	where $\mathbf{W} \in \mathbb{R}^{d_h \times d_h}$ is a learnable parameter matrix. The hidden state of $k^{th}$ fragment in passage $\mathbf{S}_k^{p}$ can be derived as similar as $S^Q$. In this paper, we consider sentence-level and clause-level as evidence fragment to support answer prediction.
	
	\paragraph{Evidence Fusion} To leverage the detected evidence, an evidence fusion layer is employed to integrate the evidence information into the hidden state of the input token sequence softly via layer normalization $\mathtt{LN}(\cdot)$ as follows: 
	\begin{equation}
		\begin{aligned}
			&\mathbf{H}_{ED} = \mathtt{LN}(\mathbf{H}_e + \mathbf{P}^{seq}\odot \mathbf{H}_e), \\&
			\mathbf{P}^{seq}_{i} =
			\begin{array}{lr}
				\mathbf{P}_{k}  & \text{if token $i$ in $k^{th}$ evidence fragment}, 
			\end{array}
		\end{aligned}
		\label{evidence_fusion}
	\end{equation}
	where the token-level evidence probability  $\mathbf{P}^{seq}$ is denoted according to the  (sentence and clause-level) evidence fragment probability distribution $\mathbf{P}_k$ ($k=1,2,...m$).

	\subsection{Evidence-Emphasized Reasoning Graph}
	\paragraph{Construction of Reasoning Graph}
	We illustrate the details about how to build the evidence-emphasized reasoning graph $\mathcal{G} = (\mathbf{V}, \mathbf{E})$ in this section. NumNet \cite{ran2019numnet} builds a directed graph with all numbers as nodes, where the direction of the edges reflects the value relationship between numbers. In contrast to NumNet, besides the number nodes $\mathbf{V}_N$, we additionally introduce sentence evidence nodes $\mathbf{V}_S$ and clause evidence nodes $\mathbf{V}_C$. i.e. $\mathbf{V} = \mathbf{V}_N \cup \mathbf{V}_{S} \cup\mathbf{V}_{C}$. The edges $E$ indicate all the relationships in the heterogeneous graph following the three situations.
	
	\begin{itemize}
		\item \textit{The edge between two number nodes}: Similar to NumNet \cite{ran2019numnet}, we also model the comparison relationship among numbers. An edge $e_{i,j}$ exists between any two number nodes $v_i$ and $v_j$, the direction of edge $e_{i,j}$ reflects the comparison relation.
		\item \textit{The edge between clause evidence nodes}: Building an edge $e_{i,j}$ between any two clause evidence nodes $e_i$ and $e_j$, if they belong to the same sentence.
		\item \textit{The edge between clause evidence nodes and sentence evidence nodes}: Building an edge $e_{i,j}$ between a clause evidence node  $e_i$ and a sentence evidence node $e_j$, if node $e_i$ belongs to sentence node $e_j$.
		\item \textit{The edge between number nodes and clause evidence nodes}: An edge $e_{i,j}$ exists between a number node $v_i$ and a clause evidence node $v_j$ when $v_i$ is in text of $v_j$.
	\end{itemize}
	
	\noindent\textbf{Initialization}  For each number node $v_i \in \mathbf{V}_N$, its representation is initialized as the corresponding token vector of $\mathbf{H}_{ED}$, i.e. $\mathbf{v}_i^{0} = \mathbf{H}_{ED}[I(v_i)]$ where $I(v_i)$ denotes the token index corresponding to node $v_i$. For nodes $v_i \in \mathbf{V}_S \cup \mathbf{V}_C$, the initial representation of $v_i$ can be derived by the weighted sum of all the corresponding tokens' representation, that is, $\mathbf{v}_i^{0} = \sum_k \alpha_k \mathbf{H}_{ED}[I(v_{ik})]$.
	%We denote all the initial node representation as $\mathbf{v}^{0} = \mathbf{v}_i \in \mathbf{V}_S \cup \mathbf{V}_C \cup \mathbf{V}_N$.
	
	\noindent\textbf{Evidence-Emphasized Information Propagation}
	To leverage the evidence to guide the reasoning over the graph, each node is assigned a weight with the probability $p_j$. We leverage relation-specific transform matrices in the message propagation to distinguish different relations among nodes. The message propagation procedure is defined as,
	\begin{equation}
		\hat{\mathbf{v}}_i^{l} = \frac{1}{|\mathcal{N}_i|} \sum_{j \in \mathcal{N}_i} p_{j} \mathbf{W}^{t_{ji}} \mathbf{v}_j^{l},
	\end{equation}
	where $\hat{\mathbf{v}}_i^{l}$ is denoted as the propagated message representation of node $v_i$ in $l^{th}$ step from its all neighbors $v_j \in \mathcal{N}_i$. $t_{ji}$ is the relations between node $v_i$ and $v_j$. $\mathbf{W}^{t_{ji}}$ is the transform matrices assigned to relation $t_{ji}$.
	
	\noindent\textbf{Updating of Node Representation}
	Formally, We update the node representation by fusing the propagated message representation of node $v_i$ obtained in last step with the information of the node as follows:
	\begin{equation}
		\mathbf{v}_i^{l+1} = \mathtt{ReLU}(\mathbf{W}_v\mathbf{v}_i^{l} + \hat{\mathbf{v}}_i^{l}),
	\end{equation}
	where $\mathbf{W}_v$ is a learnable parameter matrix. 
	
	\noindent\textbf{Fusion layer} 
	Following NumNet\cite{ran2019numnet}, we integrate the structured knowledge implied in the evidence graph into the representation of the sequence as 
	\begin{equation}
		\begin{aligned}
			\mathbf{H}_{ERG} &=[\mathbf{H}^{Q}_{ERG};\mathbf{H}^{P}_{ERG}]  =\mathtt{ResiGRU}(\mathtt{LN}(\mathbf{H}_{ED} + \mathbf{H}_{R})), \\
			\mathbf{H}_R[j] &=  \begin{array}{lr}
				\mathbf{v}_i^{l+1}  & \text{if  $j^{th}$ token in node $v_i$},
			\end{array} 
		\end{aligned}
	\end{equation}
	where $\mathtt{ResiGRU}(\cdot)$ means the composite function with residual function and a $\mathtt{GRU}$ layer.

	\subsection{Prediction Module} 
	In this section, we demonstrate the details of the answer prediction module, including five answer predictors corresponding to different answer types and an answer type predictor.
	
	\paragraph{Answer Type} 
	The answer type predictor calculates the probability distribution of different answer type choices as
	\begin{equation}
		\begin{aligned}
			&\mathbf{P}^{type} = \mathtt{softmax}(\mathtt{FFN}( \mathbf{M}^{type})), ~~
			\mathbf{M}^{type} = [\mathbf{h}^{Q};\mathbf{h}^{P}],
		\end{aligned}
	\end{equation}
	where $\mathbf{h}^{Q}$ and $\mathbf{h}^{P}$ is calculated by weighted sum with token-level representation of question and passage $\mathbf{H}^{Q}_{ERG}\in \mathtt{R}^{L_q \times d_h}$ and $\mathbf{H}^{P}_{ERG} \in \mathtt{R}^{L_p \times d_h}$.
	\paragraph{Single Span}
	Following Hu \emph{et al}.~\cite{hu2019multi}, we use a question-aware decoding strategy to predict the start and end index. Specifically,   the question representation vector, which means the summary of the question sequence information, is first computed as: 
	\begin{equation}
		\mathbf{\alpha}^{Q} = \mathtt{softmax}(\mathtt{FFN}(\mathbf{H}^Q_{ERG})),\;\; \mathbf{g}^{Q}=\mathbf{\alpha}^{Q}\mathbf{H}_{ERG}^Q.
	\end{equation}
	We leverage the pinpointed evidence to direct the prediction of the start index and end index as the following formulas: 
	\begin{equation}
		\begin{split}
			\mathbf{P}^{P}_{start},\mathbf{P}^{P}_{end} &= \mathtt{masked\_softmax}(\mathbf{P}^{seq} \odot \mathtt{FFN}(\mathbf{M})),\\
			\mathbf{M}&= [\mathbf{H}_{ERG};\mathbf{H}_{ERG} \odot \mathbf{g^Q}], \\
		\end{split}
		\label{span_prediction}
	\end{equation}
	where $\mathtt{masked\_softmax}(\cdot)$ means  $\mathtt{softmax}(\cdot)$ can only be conducted among the element not be masked, which guarantees the answer span either belongs to question or document. $\mathbf{P}^{seq}$ is the token-level evidence distribution derived  in section \ref{sec:evidence_retriever}.
	%, which is leveraged to direct the start and end index prediction.
	When some fragment is identified as evidence by the evidence detector, we enhance the probability of a token in the fragment as the start and end index, and vice versa.
	
	\paragraph{Arithmetic Expression}
	We yield the number representations $\mathbf{U} = (\mathbf{u}_{1}, \mathbf{u}_{2},..., \mathbf{u}_{N})) \in \mathbb{R}^{N\times2d_h}$ by gathering from $\mathbf{H}_{ERG}$, if $N$ numbers exists.
	Similar to Dua \emph{et al}.~\cite{dua2019drop}, we perform addition and subtraction over all numbers mentioned in the question and document context by assigning a sign (plus, minus or zero) to each number. In this way, arithmetic expression is converted into a sequence role labeling problem. The evidence information is leveraged to direct the prediction of number's sign as follows:
	\begin{equation}
		\begin{split}
			&\mathbf{P}^{sign}_i = \mathtt{softmax}(\mathbf{P}^{e}_i \odot \mathtt{FFN}( \mathbf{M}^{sign}_i)),\\
			&\mathbf{M}^{sign}_i = [\mathbf{u}_i;\mathbf{h}^{P};\mathbf{h}^{Q}], ~
			\mathbf{P}^e_{i} = (\mathbf{P}^{seq}_i, \mathbf{P}^{seq}_i, 1-\mathbf{P}^{seq}_i),\\
		\end{split}
		\label{expression_prediction}
	\end{equation}
	where $\mathbf{P}^e_{i} \in \mathbb{R}^{3}$ is a constructed weight vector for prediction of $i^{th}$ number's sign. $P^{seq}_i$ means the probability of the token as evidence at the position of the $i^{th}$ number. We increase the probability that the sign of a number is discriminate as plus and minus when the segment containing the number is identified as evidence.
	
	\paragraph{Count}
	We model count question as a 10-class classification problem (0-9). We first compute the input feature vector, integrating all the mentioned numbers, question, and passage information as follows:
	\begin{equation}
		\begin{split}
			\mathbf{P}^{count} &= \mathtt{softmax}(\mathtt{FFN}(\mathbf{M}^{count})), \;\;
			\mathbf{M}^{count} = [\mathbf{h}^{U}; \mathbf{h}^{P};\mathbf{h}^{Q}],\\
			\mathbf{h}^{U} &= \mathbf{\alpha}^{U}\mathbf{U}, \;\;\mathbf{\alpha}^{U} =\mathtt{softmax}(\mathbf{U}\mathbf{W}^{U}).\\
		\end{split}
	\end{equation}
	
	\paragraph{Multi-Spans}
	For multi-spans extraction, the probabilities are derived with a sequence role labeling method $\mathtt{SRL}(\cdot)$as the same as Segal \emph{et al}.~\cite{segal2019simple}:%Here, denote it as $\mathtt{SRL}(\cdot)$,
	\begin{equation}
		\begin{split}
			%\mathbf{M}^{sign}_i &= [\mathbf{h}^{OP};\mathbf{u}_i;\mathbf{h}^{P};\mathbf{h}^{Q};\mathbf{h}^{CLS}]\\
			\mathbf{P}^{MS} = \mathtt{SRL}(\mathbf{H}_{ERG}),
		\end{split}
	\end{equation}
	where $\textbf{P}^{MS} \in \mathbb{R}^{L\times3}$ are probability distribution of token's \texttt{BIO} tagging. 
	
	\section{Training with Distant Supervision}
	\label{sec:Data_Construction}
	%% 对于阅读理解任务来说， 推理过程是连接问题与答案之间的桥梁， 本文的出发点让机器自动去学习找到中间的桥梁，本文中是以支持证据形式存在的。
	%%然而, 通过监督学习精确获取中间信号标注的方法代价是昂贵的，我们通过简单的方式得到一般性的几条规则，远程监督学习。
	For the R-MRC task, we expect the machine can automatically learn to establish the bridge between question and answer. However, it is often expensive to obtain some intermediate signal annotations that explain the reasoning process for question answering according to the given question and passage. In this paper, we propose a few one-shot heuristic rules for evidence detection without human annotations as follows, and we utilize them as distant supervision signals to train the evidence detector.
	\begin{itemize}
		\item For an instance with span answer, we identify all the fragments (sentence-level and clause-level) that contain answer text as evidence. Moreover, the fragments are also marked as evidence if containing the topic entities in question.
		\item For an instance with arithmetic expression answer, we first find all the expressions that can obtain the answer by conducting addition or subtraction over up to 3 numbers in the context.  We identify fragments containing those numbers that can be computed to obtain answers as evidence.
	\end{itemize}

	We jointly train the evidence detector and R-MRC model in form of multi-tasks.  For evidence retrieval task, we compute the cross-entropy loss with predicted evidence label and the noised ground-truth evidence label as:
	\begin{equation}
		\mathcal{L}_{evi} = -\frac{1}{m} \sum_{k=1}^{m} y_k \log p_{k} + (1-y_k) \log(1-p_{k}),
	\end{equation}
	where $p_k$ is the probability of labeling $k^{th}$ fragment as evidence. For R-MRC model part, the probability of the answer $P(A|P;Q;E)$ can be calculated as following:
	\begin{equation}
		\begin{split}
			P(A|P;Q;E) &= \sum_{z\in \mathcal{T}}{P_z(A|P;Q;E)P(z|P;Q;E)},   
		\end{split}
	\end{equation}
	where $\mathcal{T}$ denotes all the answer types and $E$ means evidence. To train our model, we adopt the marginal likelihood objective function\cite{clark2017simple}, which sums over the probabilities of all possible annotations. The loss of answer prediction is denoted as $\mathcal{L}_{ans}$. 
	Therefore the final loss is the weighted sum of two parts of losses with hyper-parameter $\lambda$ as follows:
	\begin{equation}
		\mathcal{L} = \mathcal{L}_{ans} + \lambda \mathcal{L}_{evi}.
	\end{equation}
	During inference, we first identify the evidence via the evidence detector and then determine the answer type and the corresponding answer with the prediction module.

	\section{Experiment}
	\subsection{Dataset and Evaluation Metrics}
	We conduct experiments on an R-MRC benchmark named DROP \cite{dua2019drop} to evaluate our proposed model. DROP contains 77.4k / 9.5k / 9.6k instances for training, validation, and testing.  DROP is composed of crowd-sourced question-answer pairs based on passages from Wikipedia.  Specifically, for each question in DROP, various answer types such as date, number, or spans are involved. We take Exact Match (EM) and F1 as the evaluation metrics that are the same as previous work \cite{chen2020question,dua2019drop,ran2019numnet}. 
	
	% \begin{center}
	\begin{wraptable}{r}{0.57\textwidth}
		% \begin{table}[h]
		\small
		\centering
		\vspace{-1.83cm}
		\caption{\label{r|esults} Results on the development and test sets of DROP dataset. $\ast$ denotes our implementation results. Better results are in bold.}
		\vspace{0.3cm}
		\setlength{\tabcolsep}{1mm}
		\begin{tabular}{lcccc}
			\toprule
			\multirow{2}*{\bf Method} &\multicolumn{2}{c}{\bf Dev } &   \multicolumn{2}{c}{\bf Test}\\ 
			\cmidrule(lr){2-3} \cmidrule(lr){4-5} 
			&\bf EM &\bf F1  &  \bf EM & \bf F1 \\ 
			\midrule
			\textbf{w/o Pre-trained Model}\\
			NAQANet~\cite{dua2019drop} &46.20 &49.24 &44.07 &47.01\\
			NumNet~\cite{ran2019numnet}  &64.92 &68.31 &64.56 &67.97\\
			\midrule
			\textbf{w/ Pre-trained Model} \\
			GenBERT \cite{geva2020injecting} &68.80 & 72.30&68.6 &72.35 \\
			MTMSN~\cite{hu2019multi}  &76.68 & 80.54& 75.88&79.99 \\
			NeRd~\cite{chen2019neural}  & 78.55 & 81.85 & 78.33 & 81.71\\ 
			ALBERT-Calculator~\cite{andor2019giving} &80.22 &83.98 &79.85 &83.56\\
			% 	    NumNet+  &81.07& 84.42& 81.52& 84.84\\
			% 		QDGAT  &82.74 &85.85 &83.23 &86.38\\
			NumNet+$\ast$~\cite{ran2019numnet}   &80.74&84.09 & 80.92& 84.33\\
			QDGAT$\ast$~\cite{chen2020question}  & 82.03&85.01 &82.31 &85.65 \\
			\cdashline{1-5}[1pt/2pt]
			% 		RoBERTa&80.24&83.47&80.95&84.37\\
			% 		RoBERTa(Last layer) && &  & \\
			Baseline (RoBERTa) &80.24&83.47&80.95&84.37\\
			Baseline (ELECTRA) &81.16 &  84.22  & 81.63 &  84.98 \\
			% 		ELECTRA(Last layer) &81.72& 84.77& 82.61 &85.91 \\
			% \cdashline{1-5}[1pt/2pt]
			% 	ELECTRA(rule1)(TeacherForcing) w/ ${E}$ w/ $R$  &81.43 &84.57 & & \\
			\textbf{EviDR}  & \textbf{82.09} &\textbf{85.14} &\textbf{82.55} &\textbf{85.80} \\
			% 	Ours [(rule2)(TeacherForcing)(Solf Graph) w/ ${E}$ w/ $R$]  &&&& \\
			% 	Ours [(rule2)(NoTeacherForcing)(Hard Graph:0.1-0.9 w/ P) w/ ${E}$ w/ $R$]  & & & & \\  
			
			\midrule
			Human & & & 94.90& 96.42\\
			\bottomrule
		\end{tabular}
		\vspace{-1.0 cm}
		
	\end{wraptable}
	% \end{table}
	% \vspace{-1.2cm}
	% \end{center}

	\subsection{Experiment Settings}
	Our model is built upon the publicly available pre-trained model ELECTRA$_{large}$ \cite{clark2020electra}. We use Adam optimizer with a cosine warmup mechanism to train the model. The maximum number of epochs and batch size is set to 12 and 16, respectively.  For the parameters of ELECTRA$_{large}$, the learning rate and $L_2$ weight decay are 1.5e-5 and 0.01.  For the other parts in EviDR, they are set to 5e-5 and 5e-4. The weight $\lambda$ for loss of evidence detector is 0.2 and 0.4, which corresponds to sentence-level and clause-level evidence detection, respectively. The reasoning step over the heterogeneous graph is set to 3.
	
	\subsection{Baselines}
	\label{sec:baseline}
	We re-implemented NumNet+ \cite{ran2019numnet} \footnote{\url{https://github.com/llamazing/numnet_plus}}  and QDGAT \cite{chen2020question} \footnote{\url{https://github.com/emnlp2020qdgat/QDGAT}} as our baseline systems in this work. NumNet+ integrated relative magnitude between two numbers with a number graph and performed multi-step reasoning with graph convolution network. QDGAT\cite{chen2020question} employed a question-directed graph attention network to reasoning over a heterogeneous graph that involved entity nodes and number nodes. In addition, an pre-train model based R-MRC system with multiple answer predictors was selected as another baseline, which was denoted as Baseline (RoBERTa \cite{liu2019roberta} / ELECTRA\cite{clark2020electra}) in Table \ref{r|esults}.

	\subsection{Main Results}
	Table \ref{r|esults} displays the performance of our model and other previous competitive models on DROP. %The performance of baseline models is obtained from previous works. \cite{andor2019giving,chen2020question,chen2019neural,dua2019drop,geva2020injecting,hu2019multi,ran2019numnet}. 
	Our method achieves 82.55 EM and 85.80 F1 scores on the test set, achieving similar and even better results compared to the state-of-the-art models i.e., NumNet+ and QDGAT. Moreover, EviDR achieves 0.93 EM and 0.92 F1 score improvement over the R-MRC system denoted as Baseline in Table \ref{r|esults}. This demonstrates the effectiveness of evidence modeling.

	\subsection{Ablation Study}
	\paragraph{Effectiveness of Evidence for Reasoning}
	To analyze the effectiveness of evidence for R-MRC, ablation studies are conducted on the development set of DROP. As shown in Table \ref{Diff_answer_type}, we observe removing the evidence graph from our model,  which leads to performance declines of 0.34 EM and 0.32 F1, which indicates the effectiveness of the evidence graph. On the other hand, taking off the direction of evidence for answer prediction in Eq. \ref{span_prediction} and \ref{expression_prediction} and the evidence fusion layer in Eq.\ref{evidence_fusion}, which leads to decline in 0.36 EM and 0.46 F1. It demonstrates that integrating the evidence information into the hidden state and answer predictor facilitates the answer prediction.
	Eventually, our method achieves 0.93 EM and 0.92 F1 points improvements over the baseline system.
	
	\paragraph{Performance in Different Answer Type}
	Here, we evaluate the performance of our method on different answer types on the development of DROP. As reported in Table \ref{Diff_answer_type}, when removing the evidence graph (w/o graph) and evidence direction for answer predictor (w/o ED), performance on the number, date, and span answer decline significantly. It further demonstrates the effectiveness of our methods on different answer types.
	
	% \begin{table}[h]
	%   \small
	%   \centering
	%   \caption{\label{Diff_answer_type} Performance on different answer types in the development set of DROP. Better results are in bold. }
	%   \setlength{\tabcolsep}{2mm}{
	% 	\begin{tabular}{lcccc}
	% 	    \toprule
	% 		\multirow{2}*{\bf Method} & \multicolumn{1}{c}{\bf Number} &   \multicolumn{1}{c}{\bf\bf Date} & \multicolumn{1}{c}{\bf Span} & \multicolumn{1}{c}{\bf Overall}\\ 
	% 	   %\cmidrule(lr){2} \cmidrule(lr){3} \cmidrule(lr){4} \cmidrule(lr){5}
	% 	   \cmidrule(lr){2-2} \cmidrule(lr){3-3} \cmidrule(lr){4-4} \cmidrule(lr){5-5}
	% 		&\bf EM/F1  &  \bf EM/F1  &  \bf EM/F1   &  \bf EM/F1\\ 
	%     	\midrule
	%         ours &\textbf{}/\textbf{} & \textbf{}/\textbf{}&\textbf{}/\textbf{} & \textbf{}/\textbf{} \\
	%         ~~w/o P  &/&/&/&/ \\
	%     	~~w/o PGraph  &83.86/84.10 &53.64/61.79 & 82.65/88.12& 81.75/84.82 \\
	%     	~~w/o TForce  &{84.11}/{84.37} & {64.43}/{72.24}&{82.98}/{88.19} & {82.09}/{85.14} \\
	%     	~~w/o Loss  &{}/{} &{}/{} & {}/{} & {}/{}\\
	%     	~~w/o Evi  &83.75/83.98 & 63.33/69.52& 83.63/88.64& 82.08/85.00\\
	%     % 	~~ w/o Sent  & & & & \\
	%     % 	~~ w/o SubSent  & & & & \\
	%     	baseline &83.27/83.48    &   58.00/65.48    &    81.59/87.19   &  81.16/84.22\\
	%         % baseline &  83.85/84.06  &  58.82/65.97  &  82.14/87.72 &81.72 / 84.77  \\
	% 		\bottomrule
	% 	\end{tabular} }
	% \end{table}
	
	% \begin{center}
	\begin{table}[h]
		\small
		\centering
		\vspace{-0.6cm}
		\caption{\label{Diff_answer_type} Performance on different answer types on the development set of DROP. w/o ED means without direction of evidence for answer prediction in Eq. \ref{span_prediction} and \ref{expression_prediction} and the evidence fusion layer in Eq. \ref{evidence_fusion}. w/o Graph means removing the evidence-emphasized reasoning graph from EviDR. The Better results are in bold. }
		
		\resizebox{0.75\textwidth}{!}{
			\setlength{\tabcolsep}{2mm}{
				\begin{tabular}{lcccc} 
					\toprule
					\multirow{2}*{\bf Method} & \multicolumn{1}{c}{\bf Number} &   \multicolumn{1}{c}{\bf\bf Date} & \multicolumn{1}{c}{\bf Span} & \multicolumn{1}{c}{\bf Overall}\\ 
					%\cmidrule(lr){2} \cmidrule(lr){3} \cmidrule(lr){4} \cmidrule(lr){5}
					\cmidrule(lr){2-2} \cmidrule(lr){3-3} \cmidrule(lr){4-4} \cmidrule(lr){5-5}
					&\bf EM / F1  &  \bf EM / F1  &  \bf EM / F1   &  \bf EM / F1 \\ 
					\midrule
					EviDR &\textbf{84.11} / \textbf{84.37} & \textbf{64.43} / \textbf{72.24}&\textbf{82.98} / \textbf{88.19} & \textbf{82.09} / \textbf{85.14} \\
					~~w/o ED  &83.85 / 84.06&61.18 / 69.61&82.43 / 87.54&81.73 / 84.68 \\
					~~w/o Graph  &83.86 / 84.10 &53.64 / 61.79 & 82.65 / 88.12& 81.75 / 84.82 \\
					% 	~~w/o Evi  &83.75 / 83.98 & 63.33 / 69.52 & 83.63 / 88.64& 82.08 / 85.00\\
					Baseline &83.27 / 83.48    &   58.00 / 65.48    &    81.59 / 87.19   &  81.16 / 84.22\\
					\bottomrule
		\end{tabular} } }
		\vspace{-1cm}
	\end{table}
	%     \vspace{-1.2cm}
	% \end{center}

	\begin{wraptable}{r}{0.46\textwidth}
		\vspace{-1.1cm}
		\caption{\label{evidence_retriever} The Evaluation on the evidence detection and heuristic rules on dev dataset.}
		\vspace{0.3cm}
		\setlength{\tabcolsep}{0.6mm}{
			\begin{tabular}{lccccc}
				\toprule
				{\bf Granularity} & {\bf P} & {\bf R} & {\bf F1} & \bf{AKR}\\
				\midrule
				Sentence & 92.57 & 93.57 & 90.92 & 53.24 \\
				Clause & 89.29 & 90.08 & 86.67 & 41.31\\
				\bottomrule
		\end{tabular} } 
		\vspace{-0.6cm}
	\end{wraptable}
	\paragraph{Performance of Evidence Detector}
	To analyze whether the evidence detector in our model can correctly recognize the evidence supporting answer prediction, we evaluate the performance of the evidence detector with  Precision (\textbf{P}), Recall (\textbf{R}) and F1 as metrics. As reported in Table \ref{evidence_retriever}, for sentence-level evidence, we eventually achieve 92.57, 93.57, 90.92 on Precision, Recall, and F1 scores, respectively. And for clause-level evidence, they are 89.29, 90.08, and 86.67, respectively, which indicates the evidence detector can accurately recognize the evidence supporting answer prediction. In addition, we evaluate the heuristic rules with the metric, \underline{a}verage \underline{k}eep \underline{r}atio of sentence/clause-level evidence (\textbf{AKR}), i.e., the proportion of labeled as evidence 
	in all the fragments. As Table \ref{evidence_retriever} shown, nearly 47\% of sentences and 59\% of clauses are filtered out. It significantly reduces the redundancy and noises of evidence while ensuring the answers are available from the evidence fragments.
	% \begin{center}

	\subsection{Case Study}
	In Table \ref{case_study}, we give some examples to illustrate the effectiveness of our model compared to baseline systems. Sentence-level and clause-level evidence predicted by EviDR is in blue and bold, respectively.
	% \begin{itemize}
	%     \item The first example shows the importance of evidence for questions with number answers. NumNet+ and QDGAT fail to capture the clause "\textit{0.15\% Native American (U.S. Census)}" is the crucial evidence for answer prediction. In contrast, our model accurately recognizes all the evidence pieces,  which further facilitates the prediction of correct answers. \\ \vspace{-0.2cm}
	%     \item The second example highlights the importance of evidence for questions with span answers. We observe that NumNet+ and QDGAT only find the related text "\textit{the later category}" through the semantic matching ability. However, EviDR is capable to correctly capture what is "\textit{the latter category}" by reasoning with the detected evidence "\textit{outer provinces that were adjacent to the inner provinces and tributary states located on the border regions}" and "\textit{destruction of vientiane belonged to the later category}". \vspace{-0.2cm}
	% \end{itemize}
	The first example shows the importance of evidence for questions with number answers. NumNet+ and QDGAT fail to capture the clause "\textit{0.15\% Native American (U.S. Census)}" is the crucial evidence for answer prediction. In contrast, our model accurately recognizes all the evidence pieces,  which further facilitates the prediction of correct answers. 
	The second example highlights the importance of evidence for questions with span answers. We observe that NumNet+ and QDGAT only find the related text "\textit{the later category}" through the semantic matching ability. However, EviDR is capable to correctly capture what is "\textit{the latter category}" by reasoning with the detected evidence "\textit{outer provinces that were adjacent to the inner provinces and tributary states located on the border regions}" and "\textit{destruction of vientiane belonged to the later category}". \vspace{-0.2cm}

	% \begin{center}
	\begin{table*}[h]
		%   \small
		\scriptsize
		\centering
		\vspace{-0.4 cm}
		\caption{\label{case_study} The cases are from the development set of the DROP dataset. The predictions from the state of art models NumNet+ and QDGAT are illustrated. The last column indicates our answer prediction. Sentence-level and clause-level evidence predicted is in blue and bold, respectively.}
		%   \resizebox{1\textwidth}{!}{
		\setlength\tabcolsep{3pt}%调列距
		% 	\begin{tabular}{m{1.0in} m{3.5in} m{0.5in} m{0.5in} p{0.8in}}
		\begin{tabularx}{\textwidth}{m{0.17\textwidth}m{0.63\textwidth}m{0.08\textwidth}}
			\toprule
			\bf {Question-Answer} & \bf {Passage} & \bf Prediction \\ 
			\midrule
			\makecell[l]{\textbf{Q}: How many more \\people, in terms of \\percentage, made up \\ the \textcolor{cyan}{biggest racial} \\  \textcolor{cyan}{group} compared to \\ \textcolor{cyan}{the second smallest}? \\ \\ \textbf{A}: 97.75} & {... The population density was 73 people per square mile (28/km\u00b2).  There were 12,064 housing units at an average density of 36 per square mile (14/km\u00b2). \textcolor{blue}{\textbf{The racial makeup of the county was 97.90\% White (U.S. Census)}, 0.56\% African American (U.S. Census), \textbf{0.15\% Native American (U.S. Census)}, 0.28\% Asian (U.S. Census), 0.02\% Pacific Islander (U.S. Census), 0.36\% from Race (United States Census), and 0.74\% from two or more races.} Hispanic (U.S. Census) or Latino (U.S. Census) of any race were 0.93\% of the population. 21.3\% were of English people, 16.5\% Germans, 11.4\% Irish people, 10.7\% United States, 5.3\% danish people and 5.3\% Italian people ancestry according to Census 2000.} & \makecell[l]{\textbf{NumNet+}: \\+97.90=97.90 \\ \\ \textbf{QDGAT}: \\ +97.90=97.90 \\ \\ \textbf{EviDR}: \\+97.90-0.15=97.75} \\
			\midrule
			\makecell[l]{\textbf{Q}: \textcolor{cyan}{Southern Laos} \\ belonged to  which \\ category of territory? \\ \\\textbf{A}: tributary states} & {\textcolor{blue}{Before the Monthon reforms initiated by king Chulalongkorn, Siamese territories were divided into three categories: Inner Provinces forming the core of the kingdom, \textbf{Outer Provinces that were adjacent to the inner provinces and tributary states located on the border regions.}} \textcolor{blue}{\textbf{The area of southern Laos that came under Siamese control following the Lao rebellion  and destruction of Vientiane belonged to the later category}, maintaining relative autonomy.} Lao nobles who had received the approval of the Siamese king exercised authority on the Lao population as well as the Alak and Laven-speaking tribesmen. Larger tribal groups often raided weaker tribes abducting people and selling them into slavery at the trading hub of Champasak,. ...} & \makecell[l]{\textbf{NumNet+}: \\the later category \\ \\ \textbf{QDGAT}: \\the later category \\ \\ \textbf{EviDR}: \\tributary states} \\
			\bottomrule
		\end{tabularx}
		\vspace{-1.0cm}
		% 	}
	\end{table*}

	\section{Conclusion and Future Work}
	In this work, we propose an evidence-emphasized discrete reasoning framework named EviDR for the R-MRC task. Specifically,  we explicitly model evidence and introduce an evidence detector to recognize evidence to support answer prediction. In addition, we leverage an evidence-emphasized reasoning graph module to enhance the reasoning ability of EviDR. Experiments show that EviDR achieves remarkable performance.
	
	\section*{Acknowledgments}
	The work is supported by the National Natural Science Foundation of China (No.U1908216) and the National Key R\&D Program of China under Grant No. 2020AAA0108600.

	\bibliographystyle{splncs04}
	\bibliography{reference-1}

\end{document}